\documentclass{article}
\usepackage{spconf,amsmath,graphicx}
\usepackage{float} 
\usepackage{bm}
\usepackage{amsmath}
\usepackage{amsthm}
\usepackage{amssymb}
\usepackage{booktabs}
\usepackage{multirow}

\newcommand{\R}{\mathbb{R}}

\title{FROM SHALLOW TO DEEP: COMPOSITIONAL REASONING OVER GRAPHS FOR VISUAL QUESTION ANSWERING}

\name{Zihao Zhu$^{1}$}

\address{ 
$^1$University of Chinese Academy of Sciences}

\begin{document}

\maketitle

\begin{abstract}
In order to achieve a general visual question answering (VQA) system, it is essential to learn to answer deeper questions that require compositional reasoning on the image and external knowledge.
Meanwhile, the reasoning process should be explicit and explainable to understand the working mechanism of the model.
It is effortless for human but challenging for machines. 
In this paper, we propose a Hierarchical Graph Neural Module Network (HGNMN) that reasons over multi-layer graphs with neural modules to address the above issues.
Specifically, we first encode the image by multi-layer graphs from the visual, semantic and commonsense views since the clues that support the answer may exist in different modalities.
Our model consists of several well-designed neural modules that perform specific functions over graphs, which can be used to conduct multi-step reasoning within and between different graphs.
Compared to existing modular networks, we extend visual reasoning from one graph to more graphs.
We can explicitly trace the reasoning process according to module weights and graph attentions.
Experiments show that our model not only achieves state-of-the-art performance on the CRIC dataset but also obtains explicit and explainable reasoning procedures.

\end{abstract}


\section{Introduction}
\label{sec:intro}

\begin{figure}[t] 
\centering 
\includegraphics[width=80mm]{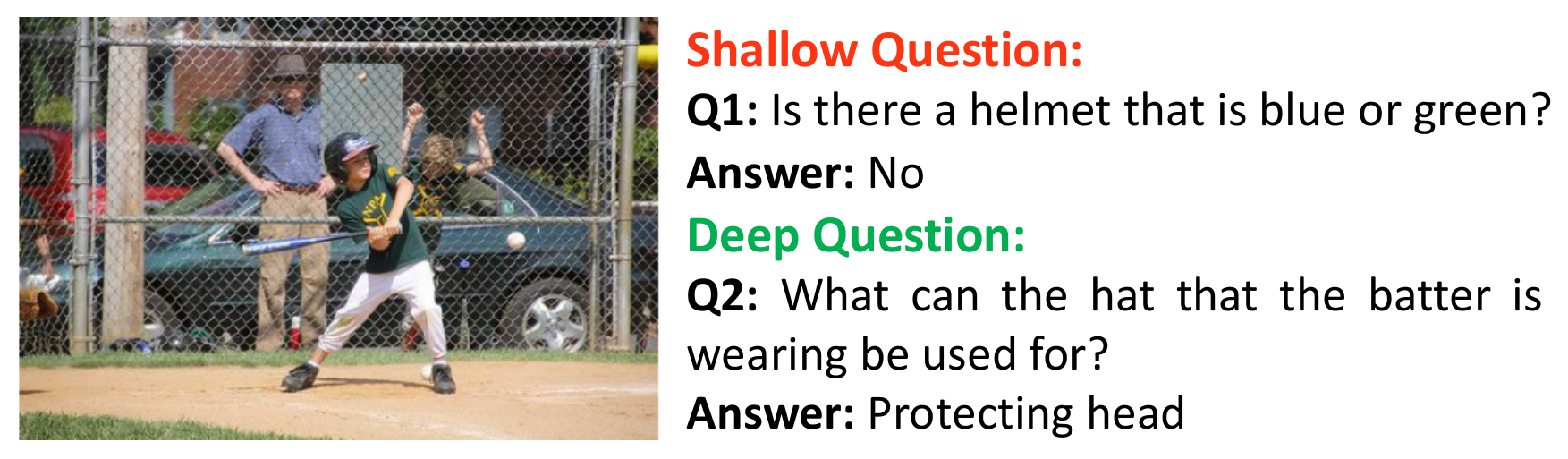} 
\caption{Examples of shallow and deep visual questions.} 
\label{fig:intro}
\vspace{-2mm}
\end{figure}

One of the goals of AI is to learn to ``see'' and ``talk'', which is consistent with Visual Question Answering (VQA) task --- aiming to answer natural language questions about an image.
Most existing works \cite{jiang2020defense,ben2019block,do2019compact} focused on ``shallow'' questions which are answerable by solely referring to the visible content of the image. 
For example, question Q1 in Fig.\ref{fig:intro} only requires recognizing the color of the helmet to answer, without multi-step reasoning incorporating external knowledge.

However, the  ``deeper'' questions (Q2 in Fig.\ref{fig:intro}) are still obstacles for these methods.
To answer this question, a desirable agent should be able to understand the semantic of the question, perceive the visual content (e.g. \texttt{helmet,boy}), incorporate commonsense knowledge (e.g. \texttt{<helmet,Used For,protect head>}) and finally compositionally reasons over these clues to predict the correct answer.
Besides, it should present the decision-making process to better understand the model's underlying working mechanism.
Therefore, how to extend shallow visual understanding to deeper compositional reasoning and meanwhile provide an explainable diagnosis are essential to achieve a long-standing general VQA goal.

Most of existing VQA models \cite{yang2016stacked,yu2019deep,yu2020cross,ma2018visual,chen2019meta,hu2018explainable} that exhibit reasoning capabilities can be divided into three categories.
Firstly, attention-based methods \cite{yang2016stacked,yu2019deep} stack attention layer that focuses on image regions relevant to the question.
The reasoning process can be post-hoc extracted by observing attention weights.
However, it is implicit reasoning process because the intermediate heat map cannot clarify what the current decision step is. 
Secondly, memory-based methods \cite{yu2020cross,ma2018visual} perform read and write operations to external memory module iteratively.
It implements reasoning by modeling interaction between multiple parts of data over several passes with attention mechanism, which still faces shortcomings of implicit reasoning.
Thirdly, module-based methods \cite{chen2019meta,hu2018explainable} implement reasoning procedure by parsing the question into a layout of sub-tasks that are carried out by separate neural sub-networks. 
The reasoning steps can be explicitly defined by each module, but relying on strong layout supervision.
Moreover, most existing methods rarely utilize commonsense knowledge that is essential for deeper reasoning.

\begin{figure*}[t]
\centering
\includegraphics[width =170mm]{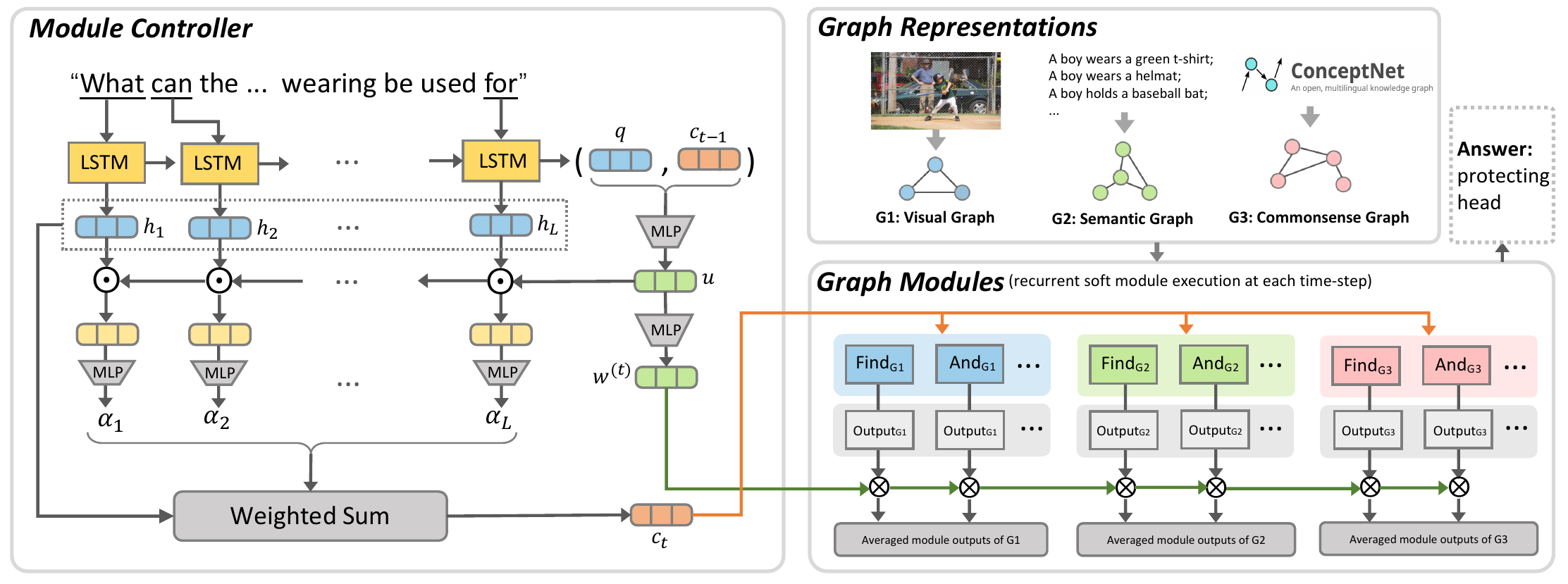}
\caption{An overview of our model at time-step $t$. It consists of graph representations, graph modules and module controller}
\label{fig:model}
\vspace{-2mm}
\end{figure*} 

In this work, we propose a Hierarchical Graph Neural Module Network (HGNMN) to address the above problems by compositional reasoning over multi-layer graphs.
Compared with holistic features, graph can model the relationships between objects and makes it easier for a model to reasons from one node to another along the edges \cite{yang2020graph}.
Therefore, we first encode an image by multi-layer graphs from the visual, semantic and commonsense views, which contain the clues supporting the answer from different modalities.
Second, we decompose the inference process into several sub-tasks based on a set of well-designed neural modules, which perform specific functions over graphs under the guidance of the semantic information of the question.
Third, we propose a module controller that supplies soft module weights and input query at each reasoning step, which makes our model fully differentiable and trainable using gradient descent without resorting to expert layouts.
Experimental results on the CRIC dataset \cite{gao2019two} demonstrate that HGNMN has comparable performance to existing modular networks and monolithic networks.
Meanwhile, the reasoning process can be explicitly traced via module weights and graph attention mechanism.

\section{Proposed Approach}	
\label{sec:format}
In this section, we elaborate on the proposed Hierarchical Graph Neural Module Network (HGNMN) for answering deeper visual questions.
Fig. \ref{fig:model} shows the overall architecture. It consists of three components:
(1) Graph Representations.
Since the clues supporting answer the question may exist in different modalities, we encode the image by constructing multi-layer graphs from the visual, semantic and commonsense views respectively;
(2) Graph Modules, a set of well-designed neural modules, that can conduct reasoning over hierarchical graphs;
(3) Module Controller guides the execution of each module at each reasoning step.

\subsection{Graph Representations}
Taking the inspiration from recent graph-based methods \cite{yu2020cross,li2019relation, zhu2020mucko} in visual and language tasks, we first encode the image as multi-layer graphs for unifying the structure representation of different modalities, including visual graph, semantic graph and commonsense graph.

\noindent\textbf{Visual Graph} represents visual objects and their relationships of the image since most of the questions in VQA are visually related.
We construct a fully-connected visual graph over a set of objects identified by Faster-RCNN \cite{ren2015faster}.
Each node corresponds to a detected object and encoded by visual feature. 
Each edge denotes the relationships between objects, encoded by relative spatial feature.

\noindent\textbf{Semantic Graph} keeps high-level abstraction of the objects and relationships within natural language also provides essential semantic information.
We first generate captions of the image with DenseCap \cite{johnson2016densecap}.
Then we construct a semantic graph using SPICE \cite{anderson2016spice}, where node represents the name or attribute of an object and edge represents the relationship between them. 
We use the averaged GloVe \cite{pennington2014glove} embeddings to represent nodes and edges.

\noindent\textbf{Commonsense Graph} contains knowledge related to the image and question that is used for answering open-domain visual questions. 
We use ConceptNet \cite{speer2017conceptnet} as original knowledge base, which can be seen as a large set of triples of the form $(h, r, t)$, where $h$ and $t$ represent head and tail entities, $r$ represents relationship between them. 
We select 10 types of relations to reduce the computation cost following \cite{gao2019two}.
A desirable knowledge retrieval should include most of the useful information while ignore the irrelevant ones.
To achieve this goal, we first use labels of visual objects as keys to extract corresponding entities of ConceptNet. 
Then we retrieve the first-order subgraph using these selected nodes from ConceptNet, which includes all edges connecting with at least one candidate node, i.e. the label is either $h$ or $t$.
However, the subgraph contains much irrelevant information.
Each object and extracted triplet is associated with probability scores, denoted as $S_l$ and $S_t$.
	We assign $a\cdot S_l+b\cdot S_t$ as the final score of the edge, where $a$ and $b$ are hyper-parameters that used to adjust the importance of the $S_l$ and $S_t$.
Then we obtain the commonsense graph by ranking and selecting top-K edges along with the nodes according to scores.
All nodes and edges are encoded by averaged GloVe embeddings\cite{pennington2014glove}.

\subsection{Graph Modules}
We first define the \texttt{Find}, \texttt{And}, \texttt{Filter} modules to attend objects that are relevant to some subjects, attributes or logic information of the question following \cite{hu2018explainable, shi2019explainable}. 
However, deeper questions commonly cover relationships (e.g. geometric positions or semantic interactions between objects) beyond mere object detection, so we propose the \texttt{Relate} module to transfer node attention maps along the edge in one graph guided by relation information of the question.
Moreover, deeper questions require collecting evidence from different modalities, so it is necessary to associate multi-layer graphs.
We additionally propose the \texttt{CrossGraph} module to extend the reasoning in one graph to more graphs.
The \texttt{Describe} module is proposed to obtain the graph features because the attentive node maps need to be transformed to an embedding to predict the answer.
Due to some questions may not require complete $T$-step reasoning, we also introduce a \texttt{NoOp} module proposed by \cite{hu2018explainable}, which can be used to pad reasoning steps to maximum length $T$.
We will describe each kind of module in detail below.
Modules of the same type only differ in inputs and parameters.

\noindent\textbf{Find [$\mathbf{X}, \mathbf{c}$]} 
is proposed to attend the relevant objects given the input query and outputs an attention map $\mathbf{a}\in \R^n$ over $n$ graph nodes.
We first transform the input query $\mathbf{c}$ and node features $\mathbf{X}$ into the same dimensions and then fuse them together to pass a MLP: 
\begin{equation}
	\mathbf{a}=\text{softmax}(f_{mlp}(F(W_1\mathbf{X},W_2\mathbf{c}))),
\end{equation}
where $W_1,W_2$ are weight matrices (as well as $W_3,\dots, W_{10}$ mentioned below), $F(\mathbf{x},\mathbf{y})= \text{ReLU}(\mathbf{x}+\mathbf{y})-(\mathbf{x}-\mathbf{y})^2$ is multimodal fusion function proposed in \cite{shi2019explainable}.

\noindent\textbf{And [$\mathbf{a}_1,\mathbf{a}_2$]} 
aims to combine attention maps $\mathbf{a}_1,\mathbf{a}_2$ generated from previous reasoning steps.
We implement it by adding two input attention weights, i.e. $\mathbf{a}= \mathbf{a_1}\oplus\mathbf{a_2}$, where $\oplus$ is element-wise addition.

\noindent\textbf{Filter [$\mathbf{a},\mathbf{c}$]} 
is proposed to find nodes relevant to the input query on the basis of the output from the previous step, which is used to further locate objects according to new input information.
We implement it based on \texttt{Find} module, i.e. $\mathbf{a}= \mathbf{a}\oplus\text{Find}(\mathbf{c})$,  where $\oplus$ is element-wise addition.

\noindent\textbf{Relate [$\mathbf{a}, \mathbf{E}, \mathbf{c}$].}
The deeper questions mostly involve the interaction between objects.
Therefore it is essential to transfer the current node to adjacent nodes via attended edge.
We first find the relevant edges given the input query. The attention weight $W_{ij}\in\mathbf{W}^{n\times n}$ of edge $e_{ij}$ is computed by:
\begin{equation}
\setlength{\abovedisplayskip}{5pt}
\setlength{\belowdisplayskip}{5pt}
	W_{ij}=\text{ReLU}(f_{mlp}(W_3\mathbf{c}\odot W_4\mathbf{e}_{ij})),
\end{equation}
where $\mathbf{e}_{ij}\in \mathbf{E}$ is edge embedding.
Thanks to graph structure, we can transfer the node weights along the attentive relations to update attention weights by matrix multiplication: $\mathbf{a}=\text{norm}(\mathbf{W}^T \mathbf{a})$,
where $\text{norm}(\cdot)$ is normalization operation.

\noindent\textbf{$\text{CrossGraph}_{G_m\to G_n}$[$\mathbf{a}_{m}, \mathbf{a}_{n}, \mathbf{X}_{m}, \mathbf{X}_{n}, \mathbf{c}$].}
Reasoning within one graph is not enough to answer deep questions.
It is essential to associate nodes of different graphs to extend the reasoning process from one graph to multi-layer graphs. 
We achieve this goal by computing attention map over nodes of graph $G_n$ guided by the node features of graph $G_m$:
\begin{align}
\setlength{\abovedisplayskip}{3pt}
\setlength{\belowdisplayskip}{3pt}
	\mathbf{a'}_n(i)=\text{softmax}(&W_7(\tanh(W_5\mathbf{X}^T_m \mathbf{a}_m+W_6\mathbf{X}_n(i))),\\
	&\mathbf{a}_n=\mathbf{a'}_n+\mathbf{a}_n,
\end{align}

where $\mathbf{X}_n(i)$ is node feature of the $i$-\textit{th} node of $G_n$.

\noindent\textbf{Describe [$\mathbf{a}, \mathbf{X}$]} 
aims to transform the attentive node features to an embedding that summarize the entire graph by computing a weighted sum of node features guided by the input attention, i.e. $\mathbf{y}=\mathbf{X}^T \mathbf{a}$.

\noindent\textbf{NoOp [a]} just outputs the input $\mathbf{a}$ without transforming, which is used to pad reasoning steps to a maximum length $T$.

\subsection{Module Controller}

To make the network fully differentiable, we draw inspire from \cite{hu2018explainable} to propose a module controller to make soft layout selection via a soft module weights distribution and supplies input query at each reasoning step $t$.
Then we execute all modules of each graph and average the outputs according to the module weights as the input for the next step.
We take one of the graphs as an example to describe the procedure in detail below.

Specifically, we first encode the question by LSTM to get all word embeddings $\{\mathbf{h}_l\}_{l=1}^L$ and question embedding $\mathbf{q}$.
At each time-step $t$, we generate intermediate embedding $\mathbf{u}$ based on question embedding and input query of previous step, i.e $\mathbf{u}=f_{mlp}([\mathbf{q};\mathbf{c}_{t-1}])$.
The input query should include question information that is more relevant to the current reasoning step.
Therefore we generate input query by textual attention over words guided by intermediate embedding:
\begin{equation}
\setlength{\abovedisplayskip}{3.5pt}
\setlength{\belowdisplayskip}{3.5pt}
	\alpha_l=\text{softmax}(f_{mlp}(\mathbf{u}\odot\mathbf{h}_l))
\end{equation}
\begin{equation}
\setlength{\abovedisplayskip}{3.5pt}
\setlength{\belowdisplayskip}{3.5pt}
	\mathbf{c}_t=\sum_{l=1}^L\alpha_l\cdot\mathbf{h}_l,
\end{equation}
where $\odot$ is element-wise multiplication.
To select which module is more important at the current step, we predict soft module weights which resemble a probability distribution over all the modules using MLP:
\begin{equation}
\setlength{\abovedisplayskip}{4pt}
\setlength{\belowdisplayskip}{4pt}
	\mathbf{w}^{(t)}=\text{softmax}(f_{mlp}(\mathbf{u}))
\end{equation}
We execute all the modules and perform a weighted average of their outputs with respect to the module weights:
\begin{equation}
\setlength{\abovedisplayskip}{5pt}
\setlength{\belowdisplayskip}{5pt}
	\mathbf{a}^{(t)}=\sum_{m\in M}w_m^{(t)}\cdot \mathbf{a}^{(t)}_m,
\end{equation}
where $w_m^{(t)}\in \mathbf{w}^{(t)}$, $M$ is module list of each graph and $\mathbf{a}^{(t)}_m$ is the output attention map of $m$-\textit{th} module.
At the final step, we extract graph features from \texttt{Describe} modules of different graphs and fuse them to predict the answer using MLP:
\begin{equation}
\setlength{\abovedisplayskip}{5pt}
\setlength{\belowdisplayskip}{5pt}
	Ans= f_{mlp}([W_8\mathbf{y}_{G_1},W_9\mathbf{y}_{G_2},W_{10}\mathbf{y}_{G_3},W_{11}\mathbf{q}])
\end{equation}

\begin{table}
\centering
\resizebox{50mm}{!}{\scriptsize
\renewcommand\arraystretch{1}
\begin{tabular}{@{}l|l|l@{}}
\toprule[1pt]
Model      & Expert layout & Acc \\ \midrule
Q-Type     & No            & 25.96    \\ 
Q-Only     & No            & 39.40    \\ 
I-Only     & No            & 14.18    \\ 
Q+I        & No            & 48.47    \\ 
ButtomUp \cite{anderson2018bottom}  & No         & 52.26    \\ 
RVC-w/o-KG \cite{gao2019two} & Yes   & 54.68    \\
RVC-$l_{ans}$ \cite{gao2019two} & No  & 51.20    \\ 
RVC \cite{gao2019two}         & Yes  & 58.38    \\\midrule
HGNMN  (full)    & No      & \textbf{60.32}    \\ \bottomrule[1pt]
\end{tabular}
}
\caption{Results on test set of CRIC dataset.}
\label{sota}
\vspace{-3mm}
\end{table}

\section{Experiments}
\subsection{Experimental Settings}
\noindent\textbf{Dataset.} 
We evaluated our model on the CRIC \cite{gao2019two}, a large VQA dataset contains 1,303,271 deeper questions than other common datasets, which are randomly split into train (60\%), validation (20\%) and test (20\%).
It also provides expert layout annotations for each question.

\noindent\textbf{Implementation Details.} For each image, we extract 36 objects along with features, labels and probability scores from Faster-RCNN \cite{ren2015faster}.
We select the top-10 captions and set $a=0.7,b=0.3$ in the graph construction.
We truncate or pad the length of question to 20 words.
The max reasoning steps $T$ is set to 12.
We adopt cross entropy loss to train the model by Adam optimizer with $0.001$ learning rate.

\subsection{Experimental Results}

\textbf{Comparison with SOTA Methods.}
In this section, we evaluate the performance of following methods on the CRIC: 
(1) \textbf{Q-Type}, \textbf{Q-Only}, \textbf{I-Only} and \textbf{Q+I} are four basic baselines proposed in \cite{gao2019two}.
(2) \textbf{BottomUp \cite{anderson2018bottom}} implements soft attention on object regions and combines the attended image features and question features to predict the answer.
(3) \textbf{RVC \cite{gao2019two}} builds upon neural module networks with supervision of expert layout.
(4) \textbf{RVC-w/o-KG \cite{gao2019two}} is a variation of RVC that doesn't use the knowledge graph to answer the question.
(5) \textbf{RVC-$l_{ans}$ \cite{gao2019two}} is a variation of RVC that can be trained without supervision of expert layout.

The results are summarized in Table \ref{sota}.
Our model achieves a new SOTA result. 
In particular, although RVC is trained with expert layout, our model still outperforms RVC by 1.94\%.
Compared with RVC-$l_{ans}$ that also does not require expert layout, our model improves by 9.12\%, which proves the effectiveness of reasoning on multi-layer graphs.

\noindent\textbf{Ablation Study.} 
We conduct ablation studies to further investigate the key components of HGNMN.
(1) We remove the visual, semantic and commonsense graph from full model respectively to analyze the contributions of each layer of graphs.
The results shown in Table \ref{table:graph} all decrease.
Thereinto, the commonsense graph is most important since most of the questions rely on external knowledge.
(2) We further evaluate the effectiveness of different neural modules in Table \ref{table:module}.
We remove the \texttt{And}, \texttt{Filter}, \texttt{Relate} and \texttt{CrossGraph} modules respectively.
The \texttt{Find}, \texttt{Describe} and \texttt{NoOp} modules are kept because answering any questions at least requires these operations.
We find that the \texttt{Relate} and \texttt{CrossGraph} modules are more important than others, since they can extend reasoning along edges and associate multiple graphs which are essential for multi-step reasoning.

\begin{table}[t]
\begin{minipage}[b]{0.4\columnwidth}
\renewcommand\arraystretch{0.99}
\centering  
\resizebox{21mm}{!}{\scriptsize
\begin{tabular}{@{}l|l@{}}
\toprule[1pt]
Model  & Acc    \\ \midrule
w/o VG & 57.63  \\
w/o SG & 59.11 \\
w/o KG & 55.25  \\ \bottomrule[1pt]
\end{tabular}
}
\caption{Ablation study of hierarchical graphs}
\label{table:graph}
\end{minipage}
\begin{minipage}[b]{0.6\columnwidth}
\renewcommand\arraystretch{0.8}
\centering
\resizebox{30mm}{!}{\scriptsize
\begin{tabular}{@{}l|l@{}}
\toprule[1pt]
Model  & Acc    \\ \midrule
w/o And & 59.12  \\
w/o Filter & 58.47 \\
w/o Relate & 57.39 \\
w/o CrossGraph & 57.14 \\ \bottomrule[1pt]
\end{tabular}  
}
\caption{Ablation study of different modules}
\label{table:module}
\end{minipage}
\vspace{-3mm}
\end{table}

\begin{figure}[t] 
\centering 
\includegraphics[width=84mm]{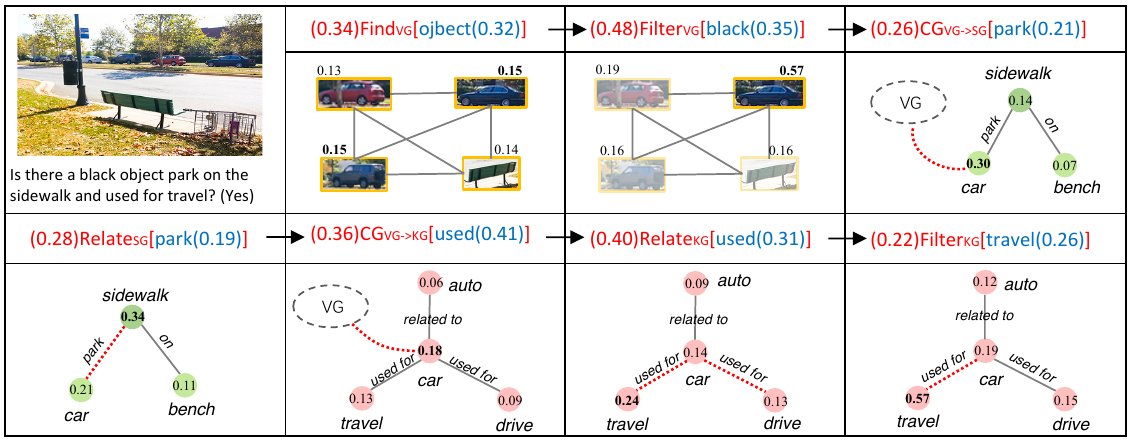}
\caption{Visualization of intermediate reasoning steps.} 
\label{fig:visual}
\vspace{-3mm}
\end{figure}
\noindent\textbf{Qualitative Evaluation}
\label{sec:visualize}
Fig. \ref{fig:visual} shows the qualitative evaluation of one example.
We visualize the results of the first 7 steps and omit the rest \texttt{NoOp} operations.
At each reasoning step, we mark the module with the largest weight in red and the most attentive word of the question in blue and present the attention map of corresponding graph.
It shows that our model can extract explicit and explainable reasoning process.
For example, it locates the ``black object'' in VG via \texttt{Filter} at the 2\textit{nd} step; finds the most relevant object in SG via \texttt{CrossGraph} at the 3\textit{rd} step; finds the nodes related to ``car'' in KG via \texttt{Relate} at the 6\textit{th} step.

\section{Conclusion}
In this paper, we propose a Hierarchical Graph Neural Module Network (HGNMN) for answering deeper questions of VQA. 
We encode the image by multi-layer graphs from different views and define a set of graph-based neural modules to extend reasoning from single graph to more graphs.
Moreover, it can be trained end-to-end without expert layout supervision.
Experimental results on CRIC dataset show that HGNMN outperforms state-of-the-art approaches.
Our model is more interpretable by observing its intermediate outputs.

\vfill\pagebreak
\clearpage

\bibliographystyle{IEEEbib}
\bibliography{refs}

\end{document}